\documentclass[runningheads]{llncs}
\usepackage{graphicx}
\usepackage{amsmath}  
\usepackage{amssymb}  
\usepackage{booktabs}  

\begin{document}

%
\title{AviationLLM: An LLM-based Knowledge System for Aviation Training
}
\author{Jia'ang Wan\inst{1} \and
Feng Shen\inst{2} \and
Fujuan Li\inst{2}\and
Yanjin Sun\inst{2}\and
Yan Li\inst{2}\and
Shiwen Zhang\inst{2}
}
\authorrunning{Wan et al.}
%
\institute{Wuhan University \and
China Eastern Technology Application Research And Development Center Co., Ltd}

%
%
\maketitle              
\begin{abstract}

Aviation training is a core link in ensuring flight safety, improving industry efficiency and promoting sustainable development. It not only involves flight simulation but also requires the learning of a great deal of professional aviation theory knowledge. In the existing training system, the knowledge is mainly imparted by the the instructors. However, the number of instructors is limited and the professional answers obtained from the Internet are not accurate enough, resulting in low training efficiency. To address this, we introduced LLM, but the basic pre-trained model cannot provide accurate answers to professional fields, so we fine-tuned it. Traditional Supervised Fine-Tuning (SFT) risk generating superficially plausible but factually incorrect responses due to insufficient data coverage. To address this, we employ Direct Preference Optimization(DPO).
This paper proposes Retrieval-Augmented LLM Alignment via Direct Preference Optimization(RALA-DPO). We select open source pre-trained LLM Qwen and adapt it to aviation theory training through DPO-based domain alignment. Simultaneously, to mitigate hallucinations caused by training data biases, knowledge obsolescence, or domain knowledge gaps, we implement Retrieval-Augmented Generation(RAG) technology that combines generative and retrieval models. RALA-DPO effectively retrieves relevant information from external knowledge bases and delivers precise and high-quality responses through the generative model. Experimental results demonstrate that RALA-DPO can improve accuracy in response to professional aviation knowledge. With integrated RAG mechanisms, this system can further improve the accuracy of answers and achieve zero-cost knowledge updates simultaneously.
\end{abstract}
\section{Introduction}

With global economic integration and increasing air transport demand, the aviation industry faces dual pressures from accelerated technological advancement and rising safety standards. Aviation theory training serves as the core foundation for aviation safety operations, providing essential theoretical support to pilots, maintenance personnel, and other professionals handling complex flight scenarios. Traditional aviation theory training is based on standardized materials, simulator-based exercises, and instructor-led knowledge transfer. However, these methods face significant limitations, including limited pedagogical resources, delayed knowledge updates, and insufficient specialized question-answering systems. These shortcomings become particularly evident when addressing contemporary challenges, such as dynamic decision making in emerging scenarios, rapidly evolving knowledge requirements, and diverse personalized learning needs. The breakthrough in the quality of the content generated by large language models (LLMs) brings opportunities for innovation in the field of aviation theory training. Using LLMs, the aviation training sector can address personnel shortages while simultaneously reducing operational costs and improving pilot training efficiency.\par
LLMs are predominantly pre-trained on large-scale general-purpose datasets. Although such broad training equips models with robust general language understanding, they underperform in tasks requiring domain-specific expertise. For example, in scientific text processing \cite{a1}, models must be capable of comprehending complex scientific terminology, concepts, and methodologies to generate accurate responses. Similarly, in e-commerce search systems \cite{a2}, mastery of domain-specific terminology is crucial for generating relevant results and recommendations. This requirement for specialized domain competency extends equally to healthcare applications \cite{a3}: large language models need to accurately understand medical terminology, diagnoses, treatment plans, and drug interactions. More notably, applications such as biomedical question answering and medical report summarization significantly rely on knowledge drawn from external sources of professional medical literature \cite{a4}. Training in aviation theory involves extensive specialized terminology and low-frequency technical vocabulary, which are underrepresented in generic training data, leading to challenges in model comprehension and application. Furthermore, LLMs trained on open-source internet data inevitably inherit biases and inaccuracies. Although effective for general tasks, such data biases may propagate incorrect or unsafe outputs in high-stakes domains like pilot training. Moreover, since most pre-training data are several years old, they cannot meet the timeliness requirements of the aviation training field. Thus, adapting general-domain pre-trained models to aviation-specific scenarios while ensuring output accuracy and timeliness remains a critical challenge.\par 
This paper proposes  RALA-DPO for constructing an aviation theory training LLM by integrating pre-trained models, DPO fine-tuning, and RAG.We constructs a DPO dataset in the aviation training domain for training large language models.Secondly, the DPO fine-tuning technique is introduced to improve the accuracy of LLM responses. Furthermore, the RAG technique is adopted to combine the fine-tuned generative model with the retrieval model, effectively realizing the retrieval of relevant information from external knowledge bases and providing accurate and high-quality answers based on the generative model. 

\section{Related work}
The field of aviation theory training imposes stringent requirements for safety, efficiency, and standardized operations. Traditional aviation training relies on standardized teaching materials, simulator training, and the imparting of human experience. Derin \cite{3} designed and implemented a simulation-based aviation English course to improve learning gains. However, when addressing decision-making in emergency scenarios, dynamic knowledge updates, and personalized learning needs, there are often problems such as limited teaching resources and outdated knowledge. In recent years, numerous studies have attempted to apply new technologies to aviation. Sihai Li \cite{4} designed an aircraft maintenance training platform based on virtual reality technology to enhance the efficiency and quality of training for maintenance personnel.

Recent advances in Large Language Models (LLMs) have established the technological foundation for their application in specialized domains. Zhang \cite{zhang2023instructfingpt} fine-tuned the models using two financial datasets, allowing them to focus primarily on financial classification tasks. Significant progress has also been made in education and interactive teaching through LLMs. For example, CyberQ \cite{agrawal2024cyberq} integrates AISecKG \cite{agrawal2023aiseckg} to combine static knowledge embedding with dynamic knowledge injection. Using structured knowledge representation and real-time updates, it generates intelligent question-and-answer outputs aligned with cybersecurity best practices. The SocraticLM model \cite{liu2024socraticlm}, fine-tuned on the SocraticTeach dataset, enhances student guidance in critical thinking and problem-solving. Additionally, the KnowGPT framework \cite{zhang2024knowgpt} converts diverse prompt templates into natural language prompts interpretable and usable by language models. This improves LLM performance in domain-specific tasks while significantly reducing API costs. Furthermore, Qwen \cite{6} achieves comprehensive advances in long-text processing and multi-task capabilities through core innovations like its Dual Chunk Attention (DCA) mechanism and Mixture-of-Experts (MoE) architecture. These abilities—including in-context learning and multimodal fusion—now offer methodological support for complex knowledge reasoning in specialized fields such as aerospace.

While pre-trained LLMs exhibit general language understanding capabilities, their outputs may not satisfy domain-specific requirements. Through fine-tuning, models can assimilate domain knowledge and adjust generation styles, thereby significantly enhancing task accuracy and consistency. Ouyang \cite{1} transformed general-purpose LLMs into instruction-following conversational agents via multi-round supervised fine-tuning (SFT) in InstructGPT. Their key innovation was constructing high-quality datasets of instruction-response pairs to align model outputs with human expectations. However, SFT’s reliance on annotated data struggles to resolve complex preference conflicts. Yao \cite{yao2024tree} proposed chain-of-thought prompting, which introduces intermediate reasoning steps to help models decompose complex tasks into manageable components. This approach enables LLMs to utilize their internal knowledge more effectively through explicit step-by-step reasoning, improving performance on tasks requiring logical inference, multi-step computation, or decision-making. OpenAI’s Reinforcement Learning from Human Feedback (RLHF) framework \cite{2} achieved value alignment through reward modeling and proximal policy optimization but faced challenges of training instability and high computational costs. Addressing this, Rafailov \cite{Rafailov2024} introduced DPO. DPO reformulates preference learning into single-stage loss function optimization via implicit reward modeling, mathematically proving equivalence to RLHF while reducing training costs by 75\%.

Meanwhile, large language models may produce erroneous outputs or hallucinations due to training data biases, knowledge obsolescence, or insufficient domain-specific knowledge. RAG technology effectively mitigates these hallucinations and knowledge latency issues through synergistic optimization between external knowledge bases and generative models. Recent advancements in RAG technology have progressed along two main trajectories: retrieval quality enhancement and knowledge-generation alignment. Representative works include HyDE \cite{Gao2023}, which improves query reformulation through hypothetical answer generation, and Atlas \cite{Izacard2023}, which employs contrastive learning to align the semantic spaces of retrievers and generators. For vertical domain applications, Self-RAG \cite{Asai2023} implements dynamic retrieval timing decisions using reflection tokens, while RA-DIT \cite{Lin2023_2} enhances domain adaptability via a two-stage adaptation process involving domain knowledge infusion followed by instruction tuning. Ren \cite{5} proposes a RAG-aided Causal Identification (RACI) model that integrates the large language model approach. However, using DPO fine-tuning or RAG alone cannot simultaneously meet the aviation theory training field's requirements for accuracy and timeliness in generated content. Therefore, this paper proposes a construction framework called RALA-DPO for an aviation theory training knowledge large model. This framework improves answer accuracy through DPO fine-tuning and prioritizes retrieving the latest authoritative documents before generating answers via RAG. This ensures that model outputs are always based on current valid knowledge, thereby guaranteeing timeliness.

\section{Background}
 The Qwen model series has made remarkable progress in architectural design, training strategies, and performance optimization, providing a strong technical foundation for fine-tuning in professional domain question-answering tasks. The Qwen2.5 \cite{6} series addresses the positional encoding bottleneck of traditional Transformer models in long-text processing through Dual Chunk Attention (DCA) technology. DCA divides long sequences into chunks and remaps relative positional relationships through a dual-chunk attention mechanism, enabling the model to efficiently handle hundreds of thousands of tokens of context. For example, the Qwen2.5-7B-Instruct-1M model only needs to be trained at a length of 32K to extrapolate to 1M length tasks and achieves nearly 100\% accuracy in complex tasks such as key retrieval. In addition, the introduction of sparse attention optimization significantly improves inference speed, with speeds increasing by 3.2-6.7 times when supporting long-text inputs, making it suitable for long-document analysis needs in professional fields.

The pre-training phase of Qwen2.5 \cite{qwen2} adopts a hierarchical optimization strategy. Firstly, it utilizes 18 trillion high-quality tokens of pre-training data (157\% more than the previous generation), covering professional fields such as mathematics and programming, and employs synthetic data generation technology to enhance data diversity. In long-text processing, the model adopts progressive context expansion training: starting with a sequence length of 4K, it gradually expands to 1M tokens, and the DCA technique reduces time complexity, solving the memory bottleneck of traditional Transformer architectures for long sequences. In the post-training phase, multi-stage reinforcement learning (DPO and GRPO algorithms) is combined to optimize capabilities such as instruction following and logical reasoning, enhancing output stability. Through strict data filtering and mixed ratio optimization, the model shows significant improvements in tasks such as commonsense reasoning and logical deduction. This feature provides a rich knowledge base for fine-tuning in professional domains. For this reason, this paper selects the Qwen model as the base model for fine-tuning to meet the requirements of aviation theory training field for generated content.

\section{Methodology}

\begin{figure}[h]
\includegraphics[width=\textwidth]{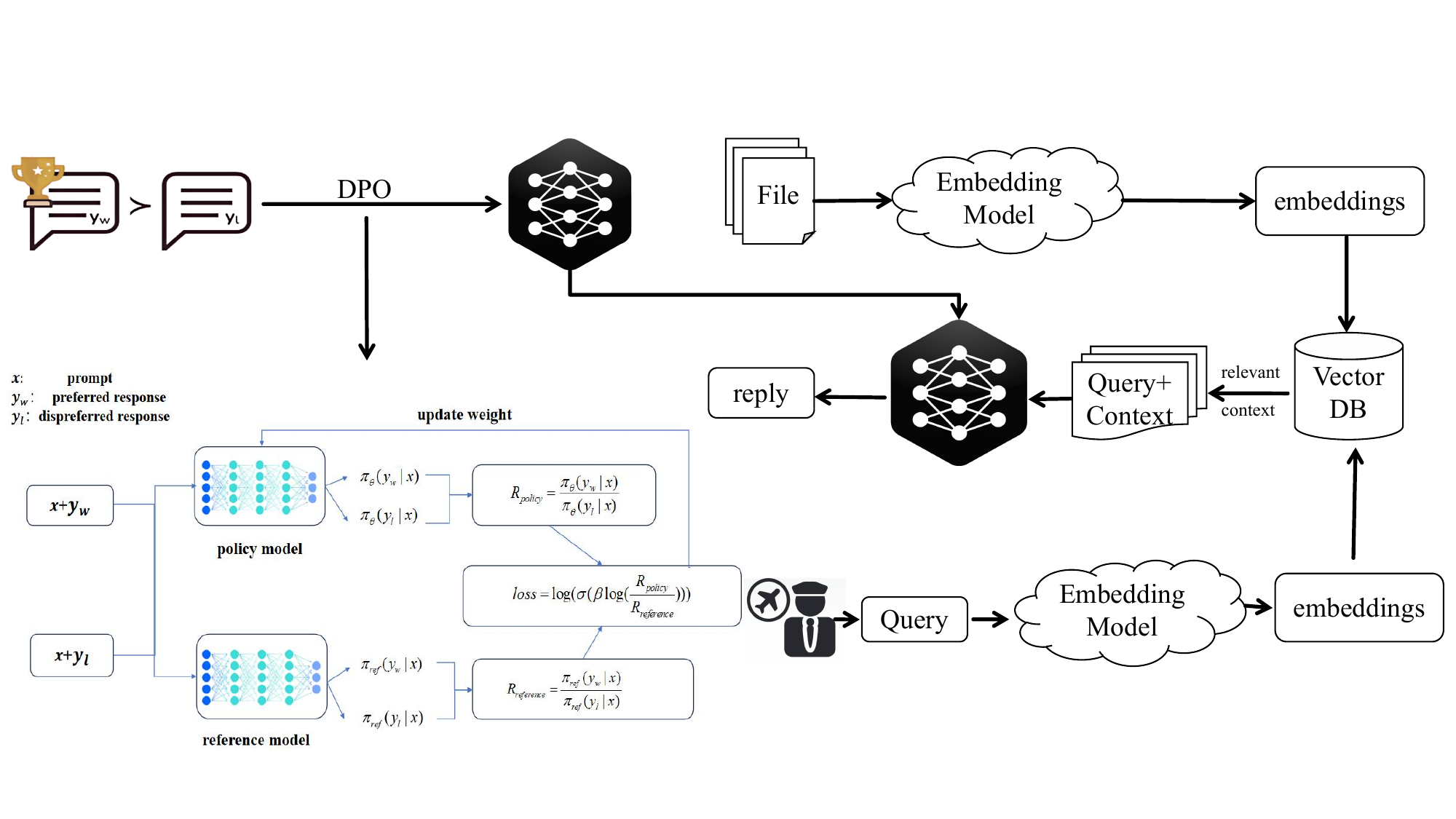}
\caption{When a user submits a query, the system first performs semantic embedding on the query to generate a corresponding semantic embedding vector. Subsequently, cosine similarity is used to calculate the vector's similarity to vectors in the vector database. This identifies the context most relevant to the query. These contexts are then substituted into a predefined prompt template. Finally, the augmented prompt, along with the original query, is input into a generative model trained using DPO to produce the reply.} \label{fig1}
\end{figure}

To address the issues of insufficient accuracy and timeliness of LLMs in the field of aviation theory training, this paper proposes RALA-DPO. We employ DPO model fine-tuning technology to refine Qwen, and integrate RAG to develop an aviation training-oriented LLM. DPO constrains the model to prefer professional responses in terms of generation preferences, ensuring the accuracy of model output, while RAG ensures the timeliness of information from the knowledge source. The collaboration of the two significantly enhances the professionalism, and timeliness of the output. The RALA-DPO is illustrated in the figure \ref{fig1}.
\subsection{Model Refinement Through Direct Preference Optimization}
This paper selects open-source pre-trained large language model Qwen and adapts it to the specialized domain of aviation industry professional training through efficient fine-tuning techniques. Specifically, we employ Direct Preference Optimization  for model refinement. DPO directly optimizes language model policies using human preference data to align model generation behaviors, while simultaneously mitigating hallucination issues in generated content. This approach enables more efficient and targeted processing of preference data, ensuring enhanced alignment with human evaluators' priorities in generated outputs. By eliminating the need for explicit reward modeling and iterative RL-based policy updates, DPO achieves superior training stability and computational efficiency compared to conventional PPO-based frameworks, while maintaining precise control over domain-specific response quality and factual accuracy required for aviation theory training scenarios.

For each input prompt $x$, we establish a pairwise comparison scenario containing two distinct outputs: a preferred response $y_w$  and a dispreferred response $y_l$. According to the Bradley-Terry model, the preference probability $P$ can be formulated as:
\begin{equation}
P(y_w \succ y_l|x) = \frac{\exp(R(x, y_w))}{\exp(R(x, y_w)) + \exp(R(x, y_l))},
\end{equation}
where the reward function $R(x,y)$ can be analytically expressed through its policy model $\pi$  and reference model $\pi_{\text{ref}}$, thereby enabling direct optimization of the policy model on preference data under proper parametrization. Specifically, the reward function can be formulated as:

\begin{equation}
R(x, y) = \beta \log\frac{\pi(y|x)}{\pi_{\text{ref}}(y|x)} + \beta \log Z(x).
\end{equation}
In the formulation, $\beta$ serves as the temperature parameter controlling the degree of policy deviation from the reference model, while $Z(x)$ acts as a normalization constant to ensure training stability. Building on this theoretical insight, the policy model can be directly optimized using human feedback data. By substituting the reward expression into the Bradley-Terry preference probability equation, we derive the simplified preference likelihood formulation:

\begin{equation}
P(y_w \succ y_l|x) = \frac{1}{1 + \exp\left(\beta\log\frac{\pi(y_l|x)}{\pi_{\text{ref}}(y_l|x)} - \beta\log\frac{\pi(y_w|x)}{\pi_{\text{ref}}(y_w|x)}\right)}.
\end{equation}
Given the constructed preference data set $ \mathcal{D} = \{(x_i, y_w^{(i)}, y_l^{(i)})\}_{i = 1}^N $, we optimize the policy parameters by maximizing the log-likelihood of observed preferences.

\begin{equation}
\mathcal{L}(\pi) = \sum_{i = 1}^N \log P(y_w^{(i)} \succ y_l^{(i)}|x_i).
\end{equation}
Substituting the simplified preference probability expression into the objective function, we derive the following DPO loss function:

\begin{equation}
\mathcal{L}(\pi)=\sum_{i = 1}^N \log\sigma\left(\beta\log\frac{\pi(y_w^{(i)}|x_i)}{\pi_{\text{ref}}(y_w^{(i)}|x_i)} - \beta\log\frac{\pi(y_l^{(i)}|x_i)}{\pi_{\text{ref}}(y_l^{(i)}|x_i)}\right),
\end{equation}
where $\sigma$ denotes the logistic sigmoid function. This maximization objective is equivalently reformulated as minimizing the negative log-likelihood:

\begin{equation}
\mathcal{L}_{DPO} = -\mathbb{E}_{(x_i,y_w,y_l) \sim D}\left[\log\sigma\left(\beta\log\frac{\pi(y_w|x_i)}{\pi_{\text{ref}}(y_w|x_i)} - \beta\log\frac{\pi(y_l|x_i)}{\pi_{\text{ref}}(y_l|x_i)}\right)\right].
\end{equation}
The gradient of the loss function simultaneously adjusts the probability distributions of both preferred responses $y_w$ and dispreferred responses $y_l$ through the following mechanism:

\begin{equation}
\nabla_{\theta}\mathcal{L} \propto -\beta\left(\frac{\pi(y_w|x)}{\pi_{\text{ref}}(y_w|x)}\nabla_{\theta}\log\pi(y_w|x) - \frac{\pi(y_l|x)}{\pi_{\text{ref}}(y_l|x)}\nabla_{\theta}\log\pi(y_l|x)\right).
\end{equation}
Through this dual-ascent optimization of the DPO loss, the model directly internalizes human preference patterns while maintaining fundamental language competencies preserved in $\pi_{\text{ref}}$ The constrained update mechanism ensures that preference alignment occurs within a trust region of the reference policy, effectively balancing three critical objectives:
Maximizing human preference satisfaction,
Preserving linguistic coherence and domain knowledge and
Minimizing hallucination through reference policy anchoring.

Through optimization of this loss function, the model directly learns to generate preferred responses through dual mechanisms of preference maximization and dispreference minimization.
\subsection{Optimization Strategies for RAG in Aviation Theory Training Domain}
Given the critical demands for timeliness in aviation theory training knowledge, this paper employs RAG technology that integrates generative models with retrieval mechanisms. The approach effectively retrieves relevant information from scalable external knowledge bases and delivers precise, high-quality responses through generative models, thereby addressing the limitations of conventional generative models in information reliability and knowledge coverage. This methodology enhances the reliability of large language model outputs and response accuracy while providing verifiable reference sources, achieving a substantial reduction in hallucination issues inherent to large language models.

First, we establish an aviation theory training knowledge base by ingesting extensive unstructured domain-specific data, including the Flight Instructor Theoretical Manual and the Basic Rules of Flight of the People's Republic of China. After these data are processed into standard text data, they are segmented into multiple knowledge fragments. A semantic embedding vector representing the semantics of each fragment is generated through a text semantic embedding model to form a collection $\mathcal{D} = \{d_1, d_2, \ldots, d_n\}$ for retrieval. The text semantic embedding model is a type of text processing model in the field of natural language processing for the specific task of text semantic extraction and compression. It compresses words, phrases, sentences, or even documents in the text into a high-dimensional embedding vector through the Transformer encoding layer. The obtained embedding vector is a compressed representation of the complete text semantics, which is easy to store and retrieve efficiently. When storing text content, this embedding vector is stored in the database together with relevant data of the original text as an index for subsequent retrieval.

During the retrieval phase, first perform semantic embedding on the aviation training-related questions raised by users to generate a semantic embedding vector $q$ for the input question. Perform relevance calculation with key documents in the knowledge library through cosine similarity calculation:
\begin{equation}
    s(q, d) = \frac{\text{Emb}_R(q) \cdot \text{Emb}_R(d)}{\parallel \text{Emb}_R(q) \parallel \cdot \parallel \text{Emb}_R(d) \parallel},
\end{equation}
where, $\text{Emb}_R(\cdot)$ is the vector representation generated by the retrieval model.Retrieve the top $k$ documents that are most relevant to query $q$ from the document collection $ \mathcal{D} = \{d_1, d_2, \ldots, d_k\}$. Then input these knowledge text fragments into the prompt template together with the question and feed them into the generation model to obtain Output $a$.
\begin{equation}
    a = \text{LM}(\text{concat}(q, D)),
\end{equation}
where, $\text{LM}(\cdot)$ is the generation model. Under the RAG framework, the language model leverages its advanced generative capabilities to synthesize contextually grounded responses by jointly analyzing both the user's query and retrieved contextual information. 

By implementing Retrieval-Augmented Generation technology, the model can generate accurate and timely content, particularly crucial for aviation professionals' theoretical training where regulatory documents frequently undergo updates. The dynamic updatability of the knowledge base enables the model to utilize the latest aviation theory training information without requiring retraining, while simultaneously mitigating hallucination issues inherent in large language models.

\section{Experiments}
In this section, we outline our experimental workflow, which includes data processing, validation of the DPO fine-tuning method, and validation of the impact of RAG technology on the models.
\subsection{Construction of the Dataset}
Given the stringent requirements for professional knowledge accuracy and timeliness in aviation theory training. This study systematically aggregates multi-domain aviation training data spanning foundational aviation theory, meteorological systems, aerodynamics, visual flight rules, and civil aviation regulations. The curated data derives from three authoritative sources: 1) certified training materials and official textbooks, 2) peer-reviewed academic literature and technical publications, and 3) regulatory documentation from the International Civil Aviation Organization (ICAO) and national aviation authorities, ensuring both provenance verification and content precision.

The data processing phase implements rigorous cleaning, structural organization, and categorical classification protocols to guarantee information integrity. Aviation knowledge texts undergo expert-guided manual annotation, constructing a premium dataset that encapsulates all critical knowledge components, resulting in 9,740 curated data pairs designated as preferred model responses. Concurrently, we generate auxiliary coarse-grained responses for identical queries using untrained large language models, establishing a dual-response framework. This comparative architecture enables systematic output optimization through response alignment analysis, ensuring strict compliance with aviation standard operating procedures. 
\subsection{Experimental Setup}
This study uses the open-source pre-trained model Qwen2.5-14B as its base model. Comparative experiments employing Supervised Fine-Tuning (SFT) and Direct Preference Optimization (DPO) techniques were conducted to evaluate their respective effectiveness. Both methods used a learning rate of 0.0003 and a batch size of 16, training them on the ATDS dataset.

For model evaluation, this paper utilizes the AI evaluator Themis-turbo from Alibaba Cloud's Bailian platform to assess two models, with the evaluation dataset derived from random sampling of ATDS. The configuration of this AI evaluator is set as follows: Models are scored across six dimensions—accuracy, relevance, completeness, source reliability, clarity of answer structure, and timeliness—using a 5-tier rating system (the more the response aligns with the scoring criteria, the higher the score).

To validate Retrieval-Augmented Generation (RAG) effectiveness, we implemented the expert evaluation method from OpsEval \cite{Liu2023}, using human experts to score output fluency, accuracy, and timeliness on a 0-5 scale. Four model variants were tested:SFT-fine-tuned Qwen2.5-14B, DPO-fine-tuned Qwen2.5-14B, SFT+RAG-enhanced Qwen2.5-14B and DPO+RAG-enhanced Qwen2.5-14B. Aviation training domain experts scored 100 sample question-answer pairs from each model configuration, with performance quality positively correlating to higher numerical ratings across all metrics.
\subsection{Experiment Results Analysis}
The comparative results of models trained with SFT and DPO, evaluated by the AI evaluator Themis-turbo, are shown in Table~\ref{tab1}, while the scores assigned by aviation theory training experts to the four models are presented in Table~\ref{tab2}.
\begin{table}[htbp]
\centering
\caption{Results of Model Comparison}
\label{tab1}
\begin{tabular}{@{} l rrrr @{}}  
\toprule
Model           & Win  & Lose & Tie & Win\% \\ 
\midrule
Qwen2.5-14B-SFT & 192  & 285  & 23  & 0.43  \\
Qwen2.5-14B-DPO & 285  & 192  & 23  & 0.57  \\
\bottomrule
\end{tabular}
\end{table}

Table~\ref{tab1} compares the performance of Qwen2.5-14B fine-tuned with DPO versus SFT on the ATDS evaluation set. The results demonstrate that the DPO-tuned Qwen2.5-14B outperforms its SFT-tuned counterpart in aviation theory training tasks. This indicates that the DPO fine-tuning technique achieves superior performance across all six evaluation dimensions—accuracy, relevance, completeness, source reliability, clarity of answer structure, and timeliness—compared to SFT fine-tuning. Evaluated by Alibaba Cloud’s Bailian platform AI evaluator Themis-turbo, the DPO-tuned model exhibits a 14\% higher win rate than the SFT-tuned model.

\begin{table}[htbp]
\centering
\caption{Expert evaluation results}  
\label{tab2}                                

\begin{tabular}{|l|l|l|l|l|}
\hline
\textbf{Model} & 
\multicolumn{3}{c|}{\textbf{Average Scores Assessed by Experts}} & 
\textbf{Total} \\
\cline{2-4}     
& \textbf{Fluency} & \textbf{Accuracy} & \textbf{Timeliness} & \\
\hline
SFT       & 2.98 & 4.38 & 3.56 & 10.92 \\
\hline
DPO       & 3.43 & 4.43 & 3.75 & 11.62 \\
\hline
SFT+RAG   & 3.98 & 4.56 & 4.42 & 12.96 \\
\hline
DPO+RAG   & 4.25 & 4.83 & 4.63 & 13.71 \\
\hline
\end{tabular}
\end{table}

In the Table~\ref{tab2}, SFT indicates SFT-fine-tuned Qwen2.5-14B, DPO indicates DPO-fine-tuned Qwen2.5-14B, SFT+RAG indicates SFT+RAG-enhanced Qwen\\2.5-14B, and DPO+RAG indicates DPO+RAG-enhanced Qwen2.5-14B. As can be seen from the table, the fluency, accuracy and timeliness of the models fine-tuned by DPO are obviously superior to those fine-tuned by SFT, which indicates that the answers generated by the models fine-tuned by DPO are more in line with the preferences of experts than those obtained by the models fine-tuned by SFT. At the same time, the model using RAG technology is significantly better than the model without RAG technology, which indicates that the use of RAG technology can enhance the understanding and generation ability of the large model to the business question and answer data in the professional field, and generate more smooth and accurate answers with stronger timeliness.

\section{Conclusion}

This paper addresses persistent challenges in aviation theory training—including instructor shortages, outdated knowledge, and domain-specific LLM hallucinations—by proposing the innovative RALA-DPO framework. The approach constructs a domain preference optimization (DPO) dataset for aviation theory training through professional data annotation and large model integration. We fine-tune the open-source Qwen model using DPO methodology to enhance accuracy in aviation theoretical training. Simultaneously, we implement Retrieval-Augmented Generation (RAG) to establish a dynamically updatable professional knowledge base, thereby improving response timeliness. RALA-DPO enhances output credibility through interpretable retrieval evidence, establishing a secure technical pathway for intelligent aviation theory training transformation. Experimental validation demonstrates that DPO significantly improves professional adaptability to aviation standard operating procedures, resolving output deviations stemming from conventional fine-tuning methods' limited data coverage. Further integration with RAG enables dynamic retrieval from updated regulatory documents and domain knowledge bases, simultaneously overcoming static training data limitations while establishing a zero-cost dynamic knowledge update mechanism.

\end{document}